\documentclass{article}
\PassOptionsToPackage{numbers, compress}{natbib}
\usepackage[final]{styling}

\usepackage[utf8]{inputenc} 
\usepackage[T1]{fontenc}    

\usepackage{url}            
\usepackage{booktabs}       
\usepackage{amsfonts}       
\usepackage{nicefrac}       
\usepackage{microtype}      
\usepackage{amsmath}
\usepackage{graphicx}
\usepackage{subfig}

\DeclareMathOperator*{\argmax}{\arg\!\max}

\title{MusicMood: Predicting the mood of music from song lyrics using machine learning}

\author{
  Sebastian Raschka\\
  Michigan State University\\
  \texttt{mail@sebastianraschka.com} \\\\
  November 2014\\
}

\begin{document}

\maketitle

\begin{abstract}
Sentiment prediction of contemporary music can have a wide-range of applications in modern society, for instance, selecting music for public institutions such as hospitals or restaurants to potentially improve the emotional well-being of personnel, patients, and customers, respectively. In this project, music recommendation system built upon on a naive Bayes classifier, trained to predict the sentiment of songs based on song lyrics alone. The experimental results show that music corresponding to a \textit{happy} mood can be detected with high precision based on text features obtained from song lyrics.
\end{abstract}

\section{Introduction}

\begin{figure}
    \centering
    \includegraphics[scale=0.40]{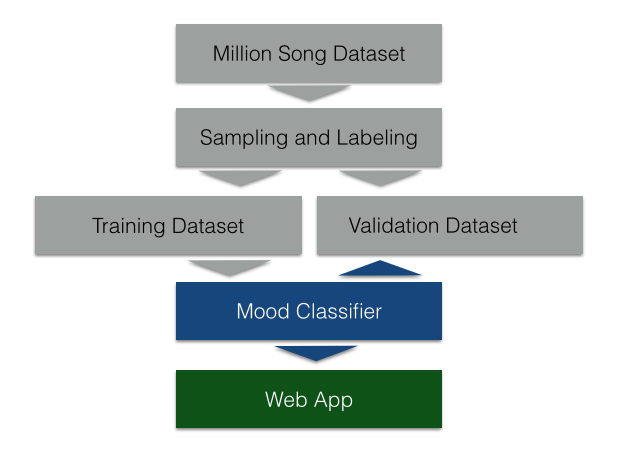}
    \caption{Flowchart summary of the MusicMood project. A subset of the Million Song Dataset \cite{hu2012genre} is divided into a training and a validation dataset. The training dataset is used to train predictive model for sentiment prediction based on song lyrics}
    \label{flowchart}
\end{figure}

With the rapid growth of digital music libraries as well as advancements in technology, music classification and recommendation has gained increased popularity in the music industry and among listeners. Many applications using machine learning algorithms have been developed to categorize music by instruments \cite{Herrera2000,Marques1999} artist similarity \cite{li2004music,schedl2012mining}, emotion \cite{lu2006automatic,kanters2009automatic,van2010automatic}, or genre \cite{tzanetakis2002musical,li2003comparative}. Psychological studies have shown that listening to music is one of the most popular activities in leisure time and that it has an enhancing effect on the social cohesion, emotional state, and mood of the listeners \cite{schafer2013psychological,vastfjall2002emotion}. The increasing number of song lyrics that are freely available on the Internet allow the effective training of machine learning algorithms to perform mood prediction and filtering for music that can be associated with positive or negative emotions. The aim of this project was to build a recommendation system that is able to predict whether a song is \textit{happy} or \textit{sad}, which can be applied to song databases in order to select music by sentiment in different social contexts (Figure \ref{flowchart}). The main contributions of this project are as follows:

\begin{enumerate}

\item Creation of a new dataset that can provide the basis of future studies on music and mood.

\item A naive Bayes classification model for mood prediction of music based on lyrics analysis.

\item An online web application to perform music mood pre-diction given artist name and song title.

\end{enumerate}

Section 2 provides a formal statement of the problem and related work. Section 3 summarizes the preprocessing and data mining steps that were conducted in this project. The experimental setup and results we obtained are presented and discussed in section 4, and the conclusions and future directions are provided in section 5.

The primary goal of this project was to build a classification model to filter for \textit{happy} music with high precision. A naive Bayes model was chosen for the lyric classification since naive Bayes classifiers are known to perform well given small sample sizes \cite{domingos1997optimality} and are successfully being used for similar binary text classification tasks such as e-mail spam detection \cite{sahami1998bayesian}. Furthermore, empirical studies have shown that the performance of naive Bayes classifiers for text categorization is comparable to Support Vector machines \cite{hassan2011comparing,go2009twitter}, while being computationally more efficient for batch and on-line learning.

The availability of open-source music datasets for research is either limited to audio feature datasets or requires manual retrieval from on-line music platforms of Creative Commons-licensed music or public domain recordings. A widely used dataset for music information retrieval (MIR) research is the freely-available Million Song Dataset \cite{msongsub} that contains audio features and metadata of a million music tracks. The musiXmatch \cite{musixmatch} dataset provides lyrics in a bag of words \cite{harris1954distributional} format for 77\% of the songs in the Million Song Dataset after application of a stemming algorithm.

While ground truth genre labels can usually be determined unambiguously through rational analysis, labeling of music by mood is a more challenging task. The perception of mood and the association of mood with different types of music are obviously subjective works. Applications of crowdsourcing approaches to collect mood ratings in Arousal-Valence (A-V) space have been designed \cite{schmidt2011modeling}, and other music mood datasets are available [21] as well; however, datasets that are providing ground truth mood labels for music are typically covering very vast and diverse sets of mood labels, which cannot be transferred to a binary categorization into \textit{happy} and \textit{sad} in an unambiguous manner.

\section{Methods}
\subsection{Data Acquisition}

A random subsample of 10,000 songs was downloaded from the Million Song Dataset \cite{msongsub} in HDF5 format. Using the provided song title and artist information from these HDF5 files, custom code was written to download the corresponding lyrics from LyricWikia \cite{lyricwikia}. Songs for which lyrics were not available --- songs that are either instrumental or not deposited in the LyricWikia database --- were removed from the dataset. The choice of acquiring the lyrics in an unprocessed format over the musiXmatch dataset was necessary for comparing different feature extraction and preprocessing steps. Custom code based on the Python NLTK library \cite{bird2006nltk} was written to identify non-English lyrics and remove these songs from the dataset using majority support based on the counts of English words vs. non-English words in the lyrics. After applying those filtering rules, the remaining dataset of 2,773 songs was randomly partitioned into a training dataset (1,000 songs) and a validation dataset (200 songs). Music labels were automatically collected from user-provided content on the music database Last.fm \cite{lastfm}. However, due to the nonexistence of mood-related tags for a majority of songs in the filtered dataset, the two mood labels (\textit{happy} and \textit{sad}) were manually assigned based on human interpretation of the lyrics and listening tests. \textit{Happy} music was defined as music that could be associated with upbeat sounds and positive themes. \textit{Sad} music was defined as music that the author related to a negative, dark, or violent theme.

\subsection{Feature Extraction}

Prior to the tokenization of the lyrics, a bag of words model \cite{harris1954distributional} (a fixed-size multiset where the order of words has no significance) was used to transform the lyrics into feature vectors. Further processing of the feature vectors include the choice of different n-gram sequences ($n \in \{1, 2, 3\}$), stop word removal based on a stop word list from the Python NLTK library \cite{bird2006nltk}, and usage of the Porter stemming algorithm \cite{porter1980algorithm} for suffix stripping. Also, different representations of the word count in the feature vectors for each song text were used, such as binarization, term frequency (\textit{tf}) computation, and term frequency-inverse document frequency (\textit{tf-idf}) computation.

The term frequency-inverse document frequency was calculated based on the normalized term frequency $\text{tf-idf}(t, d)$, which is computed as the number of occurrences of a term $t$ in a song text $d$ divided by the total number of lyrics that contain term $t$

\begin{equation}
\text{tf-idf}(t, d) = \text{tf}(t, d) \times \text{idf}(t).
\end{equation}

Let $\text{tf-idf}(t, d)$ be the normalized term frequency and $\text{idf}(t)$ be the
inverse document frequency

$$\text{idf}(t) = log{\frac{1 + n_d}{1+\text{df}(d,t)}} + 1,$$ 

where $n_d$ is the total number of lyrics and $df(d, t)$ the number
of lyrics that contain the term $t$.

\subsection{Model Selection}

Model performances using different combinations of the feature, as mentioned earlier, preprocessing techniques including hyperparameter optimization of the naive Bayes models were evaluated using grid search and 10-fold cross-validation on the 1000-song training set to optimize the F1-score. Defining the mood label \textit{happy} as the \textit{positive} class, the F1-score was computed as the harmonic mean of precision and recall

\begin{equation}
\text{F1} = 2 \times \frac{\text{precision} \times \text{recall}}{\text{precision} + \text{recall}},
\end{equation}

where

\begin{equation}
\text{precision} = \frac{\text{TP}}{\text{TP} + \text{FP}}
\end{equation}

and 

\begin{equation}
\text{recall} = \frac{\text{TP}}{\text{TP} + \text{FN}}.
\end{equation}

(TP = number of true positives, FP = number of false negatives, and FN = number of false negatives.)

Given the general notation of the posterior probability for naive Bayes classification

\begin{equation}
P(\omega_j \vert \mathbf{x_i}) = \frac  {P( \mathbf{x_i} \vert \omega_j) \times P(\omega_j)} {P(\mathbf{x_i})},
\end{equation}

the objective function in the naive Bayes model is to maximize the posterior probability given the training data where $P( \mathbf{x_i} \vert \omega_j)$ is the class-conditional probability of observing feature $\mathbf{x_i}$ belonging to class $\omega_j$:

\begin{equation}
\text{predicted class label} \leftarrow \argmax_{j = 1, \dots, m} P(\omega_j \vert \mathbf{x_i}).
\end{equation}

The class-conditional probabilities of the multi-variate Bernoulli naive Bayes model that is trained based on the binarized feature vectors are defined as

\begin{equation}
P(\mathbf{x} \vert \omega_j) = \prod_{i=1}^{m} P(x_i \vert \omega_j)^b \times (1 - P(x_i \vert \omega_j))^{1-b}.
\end{equation}

Let $\hat{P}(x_i \vert \omega_j)$ be the maximum-likelihood estimate that a particular word (or token) $x_i$ occurs in class $w_j$

\begin{equation}
\hat{P}(\mathbf{x} \vert \omega_j) = \frac{df_{x_i, y} + \alpha}{df_y + \alpha n},
\end{equation}

where $df_{x_i, y}$ is the number of lyrics in the training dataset that contain the feature $x_i$ and belong to class $w_j$. And $df_y$ is the is the number of lyrics in the training dataset that contain the feature $x_i$ and belong to class $\omega_y$. Lastly, $df_y$ is the number of lyrics in the training dataset that belong to class $w_j$, $\alpha$ is the additive smoothing parameter \cite{manning2008introduction}, and $n$ is the number of elements in the feature vector. 

Additionally, a multinomial naive Bayes model was evaluated based on the term frequencies or tf-idf, where the class- conditional probabilities are calculated as follows

\begin{equation}
\hat{P}(x_i \vert \omega_j) = \frac{\sum \text{tf}(x_i d \in \omega_j) + \alpha}{\sum N_{d \in \omega_j} + \alpha n},
\end{equation}

where $\sum N_{d \in \omega_j}$ is the sum of all term frequencies in the training dataset that belong to $\omega_j$.

For both the multi-variate Bernoulli and the multinomial naive Bayes model the class-conditional probability of encountering the song text $\mathbf{x}$ can be calculated as the product of the likelihoods of the individual terms under the \textit{naive} assumption of conditional independence between features 

\begin{equation}
P(\mathbf{x} \vert \omega_j) = P(x_1 \vert \omega_j) \times P(x_2 \vert \omega_j) \times \dots \times P(x_n \vert \omega_j) .
\end{equation}

\subsection{Software}

The Python libraries NumPy \cite{van2011numpy} and scikit-learn \cite{pedregosa2011scikit} were used for model training and model evaluation; the libraries seaborn \cite{Michael:12710} and matplotlib \cite{Hunter:2007} were used for visualization. All data, code for model training and evaluation, and the final web app have been made available at https://github.com/rasbt/musicmood.

\subsection{Experimental Setup}

After manual assignment of the mood labels and random sampling, the training dataset consisted of happy (44.6\%) and sad (55.4\%) songs; the number of happy and sad songs in the validation dataset was equal (Table \ref{distribution}). The model selection was performed via grid search and 10-fold cross- validation on the 1000-song training dataset to optimize the performance measured via F1-score. The final model was trained on the entire training dataset, the performance was evaluated on the 200-song validation dataset by measuring the receiver operating characteristic area under the curve (ROC auc), accuracy, precision, recall, and F1-score.

For initial model selection, grid search was performed on three separate naive Bayes models to select the best performing combination of feature extraction and selection approaches and parameters for each model. These three models were Multi-variate Bernoulli Bayes with binary word counts as feature vectors, multinomial Bayes with term frequency features, and multinomial naive Bayes with tf-idf features. After the three models had been individually optimized via grid search, the performance of the best performing model, from each of the three categories, was evaluated via ROC auc. The best performing model was then chosen for a more thorough optimization via grid search. During the grid search optimization, the following settings and parameters were optimized: n-gram range for tokenization, stop words removal, Porter stemming, the maximum number of features in the vocabulary (based on the $k$ most frequent tokens), a cut-off for minimum term frequency, and the $\alpha$ smoothing parameter.

\section{Results}

\begin{table}
\caption{Mood label distribution in the training and validation datasets.}\label{distribution}
\begin{center}
 \begin{tabular}{| c | c | c | c| } 
 \hline
 Mood & Training & Validation & Total \\ [0.5ex] 
 \hline\hline
 happy & 446 & 95 & 541 \\ 
 \hline
 sad & 554 & 95 & 649 \\
 \hline
\end{tabular}
\end{center}
\end{table}

The wordcloud visualizations of the most frequent words in the training dataset show an overlap between the most frequent words (\textit{love, know, come}) between the \textit{happy} and \textit{sad} songs (Figure 2). Grouping the songs by release year shows that the random 1000-song sub sample from the Million Song Dataset is bias towards more recent releases (Figure \ref{label_distribution} A); interestingly, the fraction of sad songs increases over time Figure (Figure \ref{label_distribution} B).

\begin{figure}
    \centering
    \includegraphics[scale=0.40]{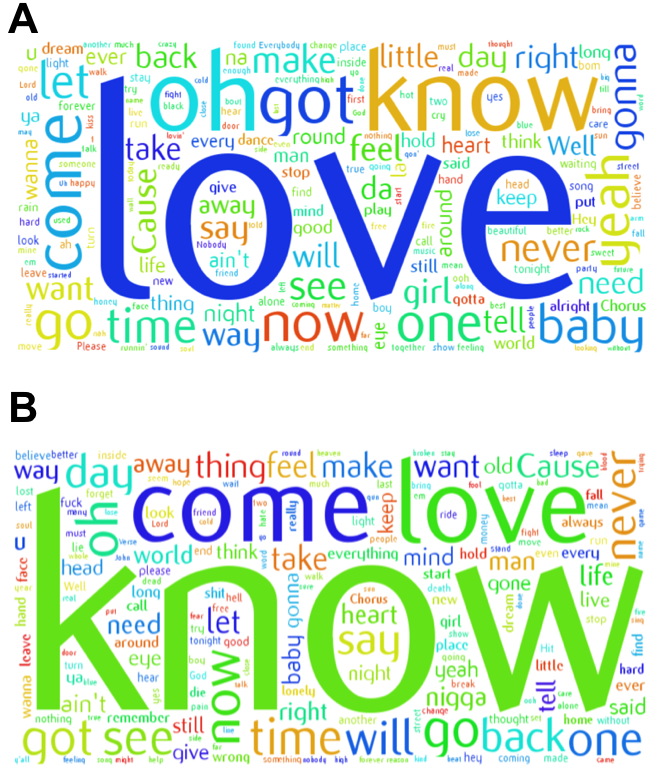}
    \caption{Wordcloud visualizations of the most frequent words of the happy songs (A) and sad songs (B) in the training dataset. The size of the words is proportional to the frequency across lyrics.}
    \label{wordclouds}
\end{figure}

After grid search for three separate naive Bayes classification models yielded an almost equal performance as shown in Figure \ref{rocs}A. The best performing model was a multinomial naive Bayes classifier (average ROC auc 0.75) with a 1-gram tf-idf feature representation after applying Porter stemming for suffix stripping and additional stop word removal. Further evaluation showed that tuning of the $\alpha$ smoothing parameter, minimum term frequency cut-off value, and maximum size of the vocabulary had little effect on the performance of the chosen classification model Figure  \ref{rocs}C-E; the attempt to increase the the n-gram range had a visibly negative effect on the classification performance \ref{rocs}F.

\begin{figure}
    \centering
    \includegraphics[scale=0.30]{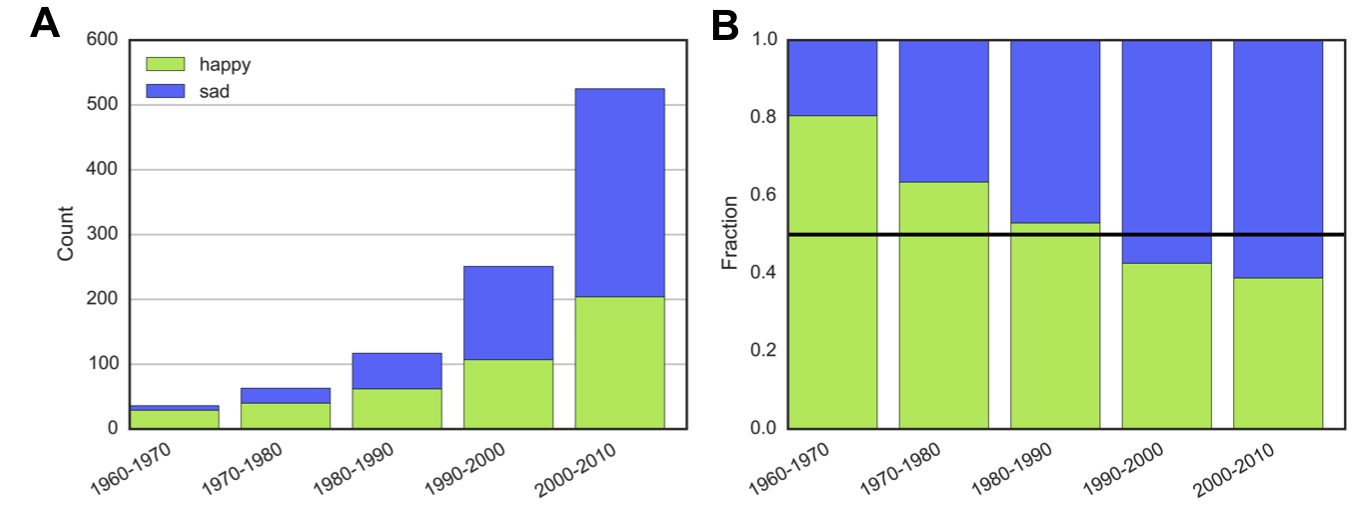}
    \caption{Distribution of \textit{happy} and \textit{sad} songs across decades in the training dataset.}
    \label{label_distribution}
\end{figure}

\begin{figure}
    \centering
    \includegraphics[scale=0.8]{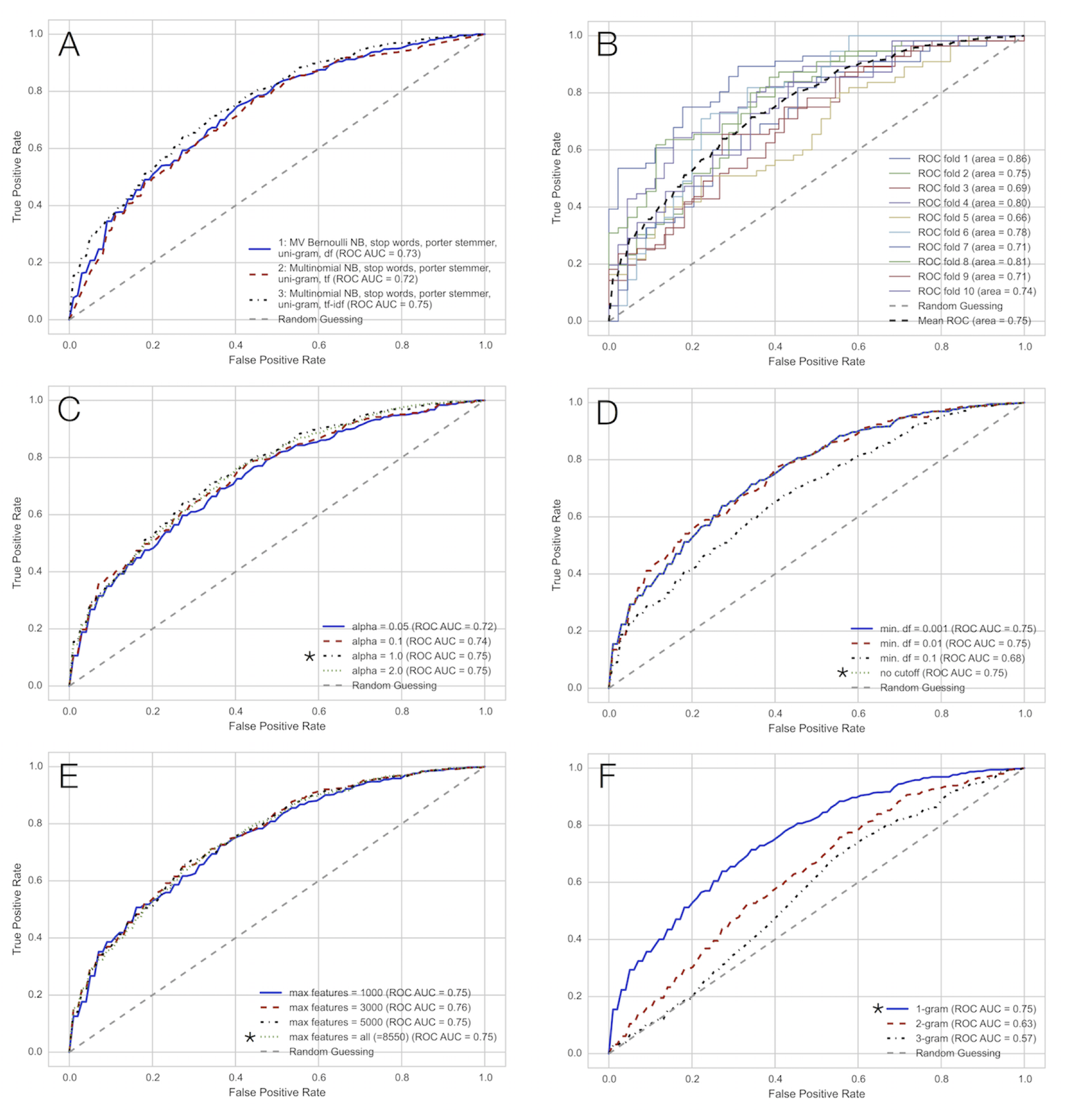}
    \caption{ROC curves of different lyrics classification models evaluated via 10-fold cross-validation on the lyrics training dataset consisting of 1,000 random songs. The true positive rate was calculated from songs labeled as \textit{happy} that were correctly classified, and the false positive rate was calculated from \textit{sad} songs that were misclassified as \textit{happy}. A: Mean ROC curves of a multi-variate Bernoulli naive Bayes classifier (1) and two Multinomial naive Bayes classifiers (2, 3) which were presented the term frequencies (tf) or term frequency-inverse document frequencies (tf-idf) of the lyrics as feature vectors. B: ROC curves of the best-selected classification model from A (3: Multinomial naive Bayes classifiers with tf-idf) with the default parameters as labeled via asterisks in C-F. C: ROC curves showing the performance of different values of hyperparameter alpha (the additive smoothing parameter). D: ROC curves comparing different thresholds for the minimum term frequency. E: ROC curves comparing the performance of the classifier for different tf-idf feature vector sizes. F: ROC curves comparing the performance of the classifier for different n-gram sizes.}
    \label{rocs}
\end{figure}

After model selection, the final classifier was trained on the complete training dataset and the performance was evaluated based on the validation dataset. The mood classifier achieved a precision performance of 99.60\% on the training set and a precision of  88.89\% on the 200-sample validation set, suggesting that it may suffer from overfitting.

\section{Discussion}

The exploratory data analysis of the training corpus showed that the fraction of sad songs increases over the years (Figure \ref{label_distribution}. However, it has to be considered that distribution of songs per year is heavily biased towards more recent releases and older music is underrepresented in the training sample. The apparent trend is still interesting and suggests that modern society could be exposed to a larger amount of \textit{sad} songs than previous generations, which makes a music recommendation system, which can be used as a mood filter, particularly interesting. All three of the different naive Bayes models that were optimized via grid search showed a better performance when stop words were removed from the lyrics (Figure \ref{rocs}). However, the higher ROC auc score of the model that was trained on tf-idf feature vectors suggests that the lyrics corpus still contained several non-relevant words that were common among both happy and sad songs as it can be seen in the wordclouds (Figure \ref{wordcloud}). As expected, the multinomial naive Bayes models showed a better performance than the Bernoulli naive Bayes models which used only binary feature vectors as input.
Although the mood classifier has a high precision for both the training (99.60\%) and validation (88.89\%) dataset, the results indicate that the cross-validation approach for model selection did not completely overcome the problem of over-fitting, which might be partially due to the large number of settings and parameters that were evaluated dur- ing grid search and the relatively small size of the training dataset. Also, the low recall rate might also be due to the equal class distribution of \textit{happy} and \textit{sad} songs in the validation dataset since the prior probabilities of the naive Bayes model were estimated from the training dataset which contained a larger fraction of sad songs. However, the high precision of the classifier is still satisfactory given the proposed goal of confidently removing \textit{sad} songs from an extensive music library before performing the genre classification. Based on the promising results, future directions include the re-evaluation of the model using a larger training dataset and mood labels selected by majority support based on labels provided by different individuals.

\section{Conclusion}

The results have shown that a naive Bayes model applied to mood classification based lyrics can predict the positive class (\textit{happy}) with high precision, which can be useful to filter a large music library for \textit{happy} music with a low false positive rate. A music library filtered in this manner could further be used as input for genre classification to filter music according to different tastes. Planned future work will include extensions to the mood classification web application to incorporate more lyrics to evaluate if the predictive performance of the classifier can be improved given a larger dataset. The extensions will include feedback about the prediction. In one extension, online learning will be implemented to update the hypothesis incrementally.

{\small
\bibliographystyle{plain}
\bibliography{references.bib}}

\end{document}